\documentclass{article}
\usepackage[utf8]{inputenc} % allow utf-8 input
\usepackage[T1]{fontenc}    % use 8-bit T1 fonts
\usepackage{PRIMEarxiv}
\let\cite\parencite

\title{A Unified Probabilistic Approach to Traffic Conflict Detection
\thanks{Officially published in \textit{Analytic Methods in Accident Research}, accessible at
\href{https://doi.org/10.1016/j.amar.2024.100369}{doi.org/10.1016/j.amar.2024.100369}}
}

\author{
  Yiru Jiao$^{1,3}$, Simeon C. Calvert$^{1,3}$, Sander van Cranenburgh$^{2,3}$, Hans van Lint$^{1}$ \\
  $^1$ Department of Transport \& Planning,\\ $^2$  Department of Engineering Systems and Services,\\ $^3$ CityAI lab,\\
  Delft University of Technology, Delft, the Netherlands\\ 
}

\begin{document}
\maketitle

\begin{abstract}
Traffic conflict detection is essential for proactive road safety by identifying potential collisions before they occur. Existing methods rely on surrogate safety measures tailored to specific interactions (e.g., car-following, side-swiping, or path-crossing) and require varying thresholds in different traffic conditions. This variation leads to inconsistencies and limited adaptability of conflict detection in evolving traffic environments, particularly as the integration of autonomous driving systems adds complexity. Consequently, there is an increasing need for consistent detection of traffic conflicts across interaction contexts.
To address this need, we propose a unified probabilistic approach in this study. The proposed approach establishes a unified framework of traffic conflict detection, where traffic conflicts are formulated as context-dependent extreme events of road user interactions. The detection of conflicts is then decomposed into a series of statistical learning tasks: representing interaction contexts, inferring proximity distributions, and assessing extreme collision risk. The unified formulation accommodates diverse hypotheses of traffic conflicts and the learning tasks enable data-driven analysis of factors such as motion states of road users, environment conditions, and participant characteristics. Jointly, this approach supports consistent and comprehensive evaluation of the collision risk emerging in road user interactions.
We demonstrate the proposed approach by experiments using real-world trajectory data. A unified metric for indicating conflicts is first trained with lane-change interactions on German highways, and then compared with existing metrics using near-crash events from the U.S. 100-Car Naturalistic Driving Study. Our results show that the unified metric provides effective collision warnings, generalises across distinct datasets and traffic environments, covers a broad range of conflict types, and captures a long-tailed distribution of conflict intensity. 
In summary, this study provides an explainable and generalisable approach that enables traffic conflict detection across varying interaction contexts. The findings highlight its potential to enhance the safety assessment of traffic infrastructures and policies, improve collision warning systems for autonomous driving, and deepen the understanding of road user behaviour in safety-critical interactions.
\end{abstract}

% keywords can be removed
\keywords{Traffic safety \and autonomous driving safety \and conflict detection \and collision risk \and Surrogate Measures of Safety}

% \newpage
%////////////////////
\section{Introduction}\label{Introduction}
%////////////////////

Collision avoidance and accident prevention are key elements in efforts to improve traffic safety \cite{Vahidi2003,Zheng2021}. Motivated by proactive prevention of accidents and not waiting for collisions to happen, Surrogate Measures of Safety were proposed as surrogates of real collisions for safety evaluation and improvement. Over the past decades, traffic conflicts have become one of the most comprehensive and prominent surrogates \cite{Zheng2014traffic,tarko2018}. A traffic conflict is defined as ``an observable situation in which two or more road users approach each other in space and time to such an extent that there is a risk of collision if their movements remain unchanged'' \cite{conflict1977}. Under this definition, every conflict is a potential collision; and every collision is a conflict until the moment when it becomes unavoidable. Despite the relative rareness when compared with safe daily interactions, successfully resolved conflicts that do not end in collisions offer opportunities to explore the emergence and resolution of collision risk \cite{tarko2018}. This has been widely recognised in the traffic safety community, and is being adopted in emerging technologies such as driving assistance and autonomous driving. 

Conflicts cannot always be directly measured due to the unclear boundary between safe and unsafe interactions, thus conflict detection often relies on surrogate metrics. These metrics are called ``conflict/safety indicators'' in the field of traffic safety \cite{Vogel2003} or ``criticality metrics'' in the field of autonomous driving \cite{junietz2018}. This difference in terminology implies different focuses on the scale of their respective research objective. Traffic safety studies concentrate more on reasoning about the causes of collisions and reducing accidents in general; while autonomous driving studies aim to ensure safe interactions between automated vehicles and other road users \cite{Razi2022}. As a result, conflict indicators are predominantly used for identifying contributing factors of collisions \cite{Saunier2011,Wu2012} and evaluating the safety of transportation infrastructure, a traffic signal system, or a specific type of interaction \cite{Saunier2007,Laureshyn2010,deCeunynck2017}. In contrast, criticality metrics are more used for predicting trajectories of other road users \cite{Wang2022}, developing autonomous driving strategies \cite{Das2021}, and assessing individual collision risk \cite{Broadhurst2005,Kim2018,Zhou2020}. In this paper, we use \textit{surrogate metrics of conflicts} to uniformly refer to both conflict/safety indicators and criticality metrics.

A variety of surrogate metrics of conflicts have been developed for different types of interactions. For instance, Deceleration Rate to Avoid Collision \cite[DRAC,][]{Cooper1976} and its variants are primarily targeted at rear-end conflicts. Time-To-Collision \cite[TTC,][]{Hayward1972,Hyden1987} along with its variants can cover both rear-end and side-swipe conflicts. Time advantage \cite{Laureshyn2010}, also known as predicted Post-Encroachment-Time (PET), is specifically tailored for path-crossing conflicts. In addition, composite indices are designed by integrating multiple metrics to deal with more complicated conflicts such as those during lane changes \cite{uno2002,Li2017}. Many summaries \cite[to name a few,][]{Arun2021,Lu2021,Wang2021,Arun2021systematic,Westhofen2022} are available for an overview of these metrics, and new metrics are being actively proposed for interactions in a two-dimensional plane \cite{Ward2015,Venthuruthiyil2022} and in traffic oscillations \cite{Kuang2015,Wang2024}. 

The diversity of these metrics entails inconsistency as collision risk is heterogeneous in different traffic conditions, between different vehicles, and is subject to changes in road user behaviour \cite{mannering2016,Zheng2021}. Some experiments show that the perception of critical TTC can vary among drivers \cite{Kusano2015} and in different traffic environments \cite{Tageldin2019,Arun2021,Chauhan2022}. For example, a TTC of 3 seconds could be dangerous for vehicles rushing on highways, but not necessarily for vehicles making a cooperative lane-change, nor for vehicles decelerating to approach an urban intersection. Similarly, a 2-second PET could be accident-prone for cars crossing their paths at an intersection, but is not uncommon for cyclists \cite{Beitel2018}. In addition, with the increasing prevalence of driving assistance systems and electric vehicles, behavioural changes in interaction may gradually influence people's perception of collision risk \cite{das2022,wessels2022}. 

Increasing efforts are devoted to overcoming these inconsistencies. For example, there is a growing trend towards combining multiple metrics \cite[e.g., ][]{Kuang2015,Nadimi2021,Mazaheri2023}, particularly by applying deep learning methods \cite{Formosa2020,AbdelAty2023}. Such a combination builds understanding from existing metrics, and thus remains constrained by their underlying assumptions of conflicts. Many studies have also looked at determining robust thresholds to distinguish unsafe conflicts from safe interactions. If there is ground truth, the threshold should optimise the accuracy of conflict identification for, e.g., issuing crash warnings \cite{Das2021}. When no label is available, the threshold selection often follows heuristic rules \cite{Panou2018,Das2020,Nadimi2022}, or is guided by the extreme value theory modelling to satisfy certain hypotheses \cite{Farah2017,Borsos2020,Niu2024}. 

However, as the development of autonomous driving rapidly advances, the inconsistency in conflict detection poses new challenges for traffic safety. Automated vehicles require a unified approach to estimating collision risk as they navigate in various road environments. Without consistent estimation, these vehicles may struggle to reliably interpret and respond to traffic conflicts, thereby jeopardising road safety. As different levels of vehicle automation increasingly share the road, traffic conditions will become more complex. Road user interactions will no longer be solely between humans, but will also involve driving algorithms developed by different manufacturers, creating new layers of complexity. Consequently, in the long run, the assessment and improvement of traffic safety will become more challenging.

To address the challenges, this study introduces a new approach for consistent and comprehensive conflict detection. First, we propose a unified framework of traffic conflict detection by formulating Bayesian collision risk. Second, we present a series of statistical learning tasks that apply the theoretical framework to practical conflict detection. Our approach considers a traffic conflict as an extreme event of normal interactions, and quantifies its context-dependent and proximity-characterised collision risk. Then the learning tasks break down conflict detection into interaction context representation, proximity distribution inference, and extreme event assessment.  Theoretically, any existing surrogate metric of conflicts is a special case under the framework. With this approach, conflict detection can be consistent across different traffic environments, and traffic safety evaluation can involve more comprehensive considerations.

The rest of this paper is organised as follows. Section \ref{sec: theory} first explains the unified framework of traffic conflict detection. Then Section \ref{sec: learning} presents the statistical learning tasks to apply conflict detection, including probability estimation and conflict intensity evaluation. In Section \ref{sec: demonstration}, demonstration experiments are designed, performed, and analysed to show the performance of the approach. Finally, Section \ref{sec: conclusion} concludes this paper and envisions future research.

%//////////////////////
\section{A unified framework of traffic conflict detection}\label{sec: theory}
%//////////////////////
This section introduces our unified framework of traffic conflict detection. We frame conflict detection as quantifying the risk of a potential collision $c$ between two or more road users, based on their proximity $s$ and other contextual observables $X$, i.e., $p(c|s,X)$. This risk quantification uses Bayes' theorem to consider the interaction situation of an event, the typical proximity behaviour of road users in the interaction context, and the probability variation with conflict intensity. These considerations are summarised in Equation \eqref{eq: theory conflict detection} for a preliminary overview, where the symbols are defined in Table \ref{tab: symbols}. 
\begin{equation}\label{eq: theory conflict detection}
    p(c|s,X)=\iint p(c|s,\phi)p(\phi|\theta)p(\theta|X)\mathrm{d}\phi\mathrm{d}\theta \text{, where } p(c|s,\phi):=C(n;s,\phi)
\end{equation}

\begin{table}[h]
\centering
\caption{Symbols and their definitions in the unified framework of traffic conflict detection.}
\label{tab: symbols}
\begin{tabular}{cl}
\toprule
\textbf{Symbol} & \textbf{Definition}\\ \midrule
$c$ & A potential collision, i.e., conflict\\
$s$ & Proximity, the spatial-temporal closeness between road users\\
$X$ & Observables that can be measured to describe interaction situations\\
$\theta$ & Representation for the interaction context involving key information selected from $X$\\
$\phi$ & Parameters used to characterise the proximity distribution of road users in a given $\theta$\\
$n$ & Conflict intensity, a conflict at intensity $n$ occurs once per $n$ times in the same interactions\\
$C$ & Probability of a conflict with intensity $n$ at proximity $s$ in the context characterised by $\phi$ \\ \bottomrule
\end{tabular}
\end{table}

The Bayesian components of the integral in Equation \eqref{eq: theory conflict detection} represent a probabilistic breakdown of collision risk. Starting from the outermost component, the first term $p(\theta|X)$ denotes the probability of a specific interaction context $\theta$ given observable measurements $X$, such as road user motion states and environmental conditions. The second term $p(\phi|\theta)$ describes the probability of the parameters $\phi$ that characterise proximity patterns conditioned on the interaction context $\theta$. Finally, the innermost term $p(c|s,\phi)$ expresses the probability of a potential collision occurring at a proximity $s$, based on the proximity distribution parameterised by $\phi$ in the specific context. This probability also accounts for conflict intensity $n$, which reflects the extremeness extent of the conflict within a spectrum of interactions in the same context.

The following subsections provide more detailed explanations and derivations to establish the framework. We begin with defining the probabilistic collision risk that depends on proximity and other situational variables. Then we assume a conflict hierarchy of risk perceptions and reactions in an interaction context, and describe the conflict hierarchy quantitatively by measuring how road user behaviour varies with proximity. Lastly, we consider conflicts as extreme events within a spectrum of interactions, and use extreme value theory to relate conflict probability with conflict intensity. 

%---------------------
\subsection{Context-dependent collision risk}\label{sec: context-dependent collision risk}
%---------------------
In this subsection, we first present a general probabilistic description of conflict detection which depends on the context of interaction. A major branch of surrogate metrics of conflicts is based on proximity. Proximity means physically less space and time for reactions to prevent a potential collision; and due to the physical constraints, people perceive increased risk when the distance to an approaching object is closer \cite{Schiff1979,Teigen2005}. Therefore, a proximity-based conflict metric can cover both the objective aspect of collision risk and people's subjective perception of the collision risk. 

Proximity-based conflict detection can be generally formulated as $p(c|s, X)$, which estimates the probability of a potential collision $c$ based on proximity $s$ ($s\geq 0$) and other observables $X$ of the interaction context. The probabilistic formulation enables consistent evaluation across contexts, as probability has a normalised range between 0 and 1. This consideration has also been taken by some studies, for example, \textcite{Saunier2008} and \textcite{deGelder2023}. For an interaction involving two or more road users, $s$ is a measure of the spatial-temporal closeness between the road users; and $X$ can include, but is not limited to, motion states of the road users, traffic states, road layouts, and weather conditions.

Not all of the information in $X$ is used, nor might it all be useful. Every existing surrogate metric of conflicts selects some key information, i.e., variables, and makes a hypothesis based on the selected information. For example, TTC uses the relative velocity between two approaching road users and assumes no movement change (constant acceleration) at the moment of conflict; DRAC also uses the relative speed but assumes an immediate brake. Here we use $\theta=\{\theta_1,\theta_2,\dots,\theta_k\}$ to denote $k$ variables extracted from $X$, and $\theta$ are assumed to adequately represent the interaction context compressed out of the situation $X$. Considering all possible selections of $\theta$, Equation \eqref{eq: conflict detection with context integrated} holds according to Bayes' theorem.
\begin{equation}\label{eq: conflict detection with context integrated}
    p(c|s,X)=\int p(c|s,\theta)p(\theta|X)\mathrm{d}\theta
\end{equation}
If the selection of key information is deterministic as $\theta=\mathbb{R}(X)$, where $\mathbb{R}$ refers to representation learning of the interaction context, $p(\theta|X)$ in Equation \eqref{eq: conflict detection with context integrated} becomes a Dirac delta function of $\delta(\theta-\mathbb{R}(X))$. A Dirac delta function $\delta(x)$ has a value of 1 when $x=0$ and 0 for any other values of $x$. Therefore, Equation \eqref{eq: conflict detection with context integrated} can be approximated to Equation \eqref{eq: conflict detection with context}.
\begin{equation}\label{eq: conflict detection with context}
    p(c|s,X)=p(c|s,\theta)=p(c|s,\mathbb{R}(X)) \text{, if } p(\theta|X)=\delta(\theta-\mathbb{R}(X))
\end{equation}

\begin{figure}[hbt]
    \centering
    \begin{subfigure}[b]{0.4\textwidth}
        \centering
        \includegraphics[width=\textwidth]{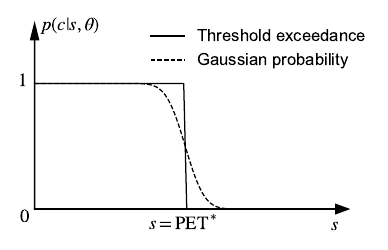}
        \caption{PET: Post-Encroachment Time.}
        \label{fig:pet_example}
    \end{subfigure}
    \begin{subfigure}[b]{0.4\textwidth}
        \centering
        \includegraphics[width=\textwidth]{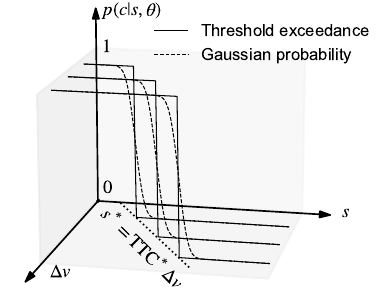}
        \caption{TTC: Time-to-Collision.}
        \label{fig:ttc_example}
    \end{subfigure}
    \caption{Illustration examples of context-dependent proximity-based conflict probability.}
    \label{fig:examples}
\end{figure}

To better explain how a surrogate metric of conflicts fits into our formulation, we provide two examples. The first is Post-Encroachment Time (PET), where $s$ is the time interval between one vehicle leaving a conflict area and another vehicle arriving in the same conflict area. The $\theta$ for PET is the existence of a conflict area. As shown in Figure \ref{fig:examples}(a), $p(c|s,\theta)$ can be a step function with probability 1 when PET exceeds a threshold PET$^*$ and 0 otherwise; or a continuous function based on a cumulative Gaussian probability that increases gradually from 0 to 1 and is 0.5 at the threshold PET$^*$. Then another example uses $\text{TTC}=s/\Delta v$ as shown in Figure \ref{fig:examples}(b). For two vehicles approaching each other, $s$ is the net distance between them; while $\Delta v$ is their relative speed, and serves as a univariate context $\theta$. The probability $p(c|s,\theta)$ varies at different relative speeds. If there is a critical threshold $\text{TTC}^*$ that differentiates safe and unsafe interactions, the threshold of proximity then follows $s^*=\text{TTC}^*\Delta v$. 

%---------------------
\subsection{Proximity-characterised conflict hierarchy}\label{sec: proximity-characterised conflict hierarchy}
%---------------------
Proximity patterns in an interaction context reflect the aggregated behaviour of road users in this context, which is shaped by people's perception of collision risk. Within the same context of interaction, a shorter spatial or temporal gap between road users consistently implies more risk of collision. This increase in collision risk with decreasing proximity motivates road users to maintain an acceptable distance from the others at varying levels \cite{Camara2020}. Therefore, the proximity behaviour of road users embodies a conflict hierarchy perceived by the road users. On the one hand, proximity pushes or prevents the road users from approaching each other; on the other hand, the road users adjust their behaviours to maintain comfortable proximity. 

We characterise the proximity behaviour of road users with a conditional probability distribution. Seeing proximity as a random variable $S$, in a certain interaction context $\theta$, the conditional probability distribution of $S$ given a context $\theta$ is $p(s|\theta)$. We use $f_S(s|\theta;\phi)$ to describe the density function of $p(s|\theta)$, where $\phi$ is a set of parameters. To incorporate context-dependent proximity behaviour into conflict detection, we consider all possible sets of $\phi$ in a context $\theta$, thus integrate $p(\phi|\theta)$ in Equation \eqref{eq: conflict detection with proximity dist integrated} according to Bayes' theorem. 
\begin{equation}\label{eq: conflict detection with proximity dist integrated}
    p(c|s,\theta)=\int p(c|s,\phi)p(\phi|\theta)\mathrm{d}\phi
\end{equation}
As we parameterise the conditional probability $p(s|\theta)$ with $f_S(s|\theta;\phi)$, inferring $p(s|\theta)$ is to obtain $\phi$, and we denote this inference as $\phi=\mathbb{I}(\theta)$. When the inference is deterministic, $p(\phi|\theta)$ in Equation \eqref{eq: conflict detection with proximity dist integrated} becomes a Dirac delta function $\delta(\phi-\mathbb{I}(\theta))$. In the same way as explained when deriving Equation \eqref{eq: conflict detection with context}, we can derive Equation \eqref{eq: conflict detection phi}.
\begin{equation}\label{eq: conflict detection phi}
    p(c|s,\theta)=p(c|s,\phi)=p(c|s,\mathbb{I}(\theta)) \text{, if } p(\phi|\theta)=\delta(\phi-\mathbb{I}(\theta))
\end{equation}

Numerous empirical studies have observed the transition from comfort to discomfort when a road user is approached by other vehicles \cite{Taieb2001,LewisEvans2010,Siebert2014,Siebert2017}. Therefore, we can generally assume that $p(c|s,\phi)$ monotonically increases while $s$ is decreasing. In a conflict hierarchy characterised by $\phi$, the closer the proximity between approaching road users, the less safe they are and the higher the probability of a potential collision. This monotonicity of $p(c|s,\phi)$ implies that Equation \eqref{eq: probability limit} holds for any $\phi$. 
\begin{equation}\label{eq: probability limit}
\begin{cases}
\begin{aligned}
    \lim_{s\rightarrow\infty} p(c|s,\phi)&= 0\\
    \lim_{s\rightarrow 0} p(c|s,\phi)&= 1
\end{aligned}    
\end{cases}
\end{equation}

\begin{figure}[htb]
    \centering
    \includegraphics[width=0.88\textwidth]{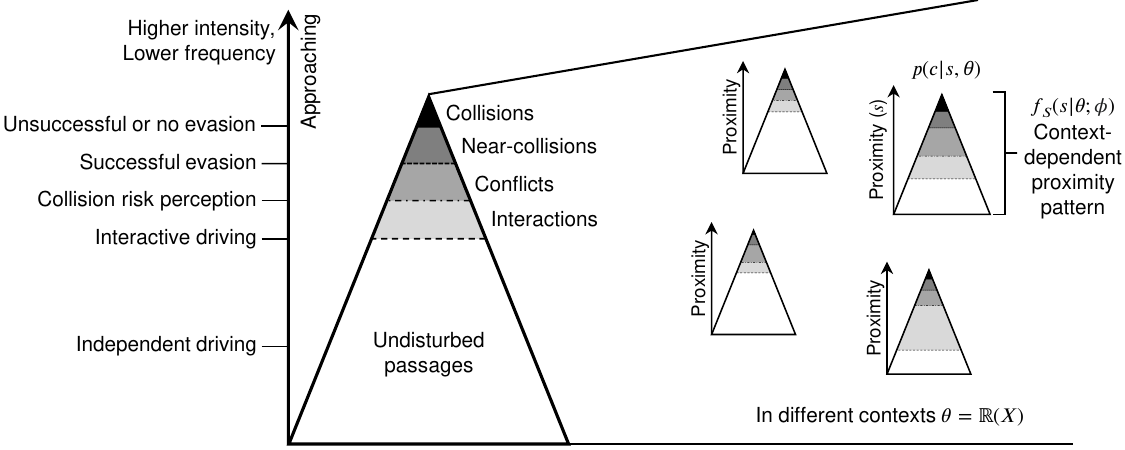}
    \caption{Proximity-characterised conflict hierarchy varies in different interaction contexts.}
    \label{fig: safety_pyramid}
\end{figure}

We use Figure \ref{fig: safety_pyramid} to help explain the proximity-characterised conflict hierarchy. On the left of Figure \ref{fig: safety_pyramid}, we present an adapted pyramid of conflict hierarchy \cite{Hyden1987}, which conceptually relates conflict intensity and frequency, and reflects road user behaviour in balancing safety and efficiency \cite{Laureshyn2010}. When the approaching between road users entails a potential collision, the closer they are, the higher the conflict intensity and the lower the frequency of such conflicts. When the proximity is too small for safe interaction, a conflict emerges entailing a potential collision. Furthermore, an accident will happen if there is no (successful) evasion or the proximity is too small to prevent a collision. This hierarchy of conflicts is not fixed, but has varying shapes in different interaction contexts. As illustrated in the right of Figure \ref{fig: safety_pyramid}, $f_S(s|\theta;\phi)$ is designed to capture such context-dependent proximity patterns. Consequently, the estimated probability $p(c|s,\phi=\mathbb{I}(\theta))$ may vary for the same proximity in different contexts, and evolve as the interaction context changes over time.

%---------------------
\subsection{Extreme value theory-based interaction spectrum}\label{sec: EVT-based interaction spetrum}
%---------------------
Traffic interactions exist on a spectrum varying in conflict intensity and potential consequences. Within this spectrum, we assume that conflicts are extreme events at different levels of collision risk and collisions are the most extreme cases. Then we can establish the relation between conflict intensity and conflict probability utilising extreme value theory. In traffic safety research, the extreme value modelling has been traditionally used to treat crashes as the extreme events of conflicts and extrapolate average crash risk from traffic conflicts \cite{Songchitruksa2006,Zheng2014freeway,Farah2017,Tarko2021,Nazir2024,Niu2024}. Here we zoom out the view and provide a derivation that collisions and conflicts are both extreme events of ordinary interactions, at varying levels of intensity.

Our derivation uses a similar logic to lifetime survival analysis, which is a branch of statistics to model the expected duration of time until one event, such as deaths or failures of medical treatment,  occurs \cite{reiss1997}. The essential idea of lifetime survival analysis is to estimate the probability of events from the records of human lives or running systems, where the frequency of these events reduces when living time increases. Therefore, the survival function is $S(t)=\Pr(T>t)$, representing the cumulative probability that the duration of survival $T$ is longer than some specified time $t$. The longer the $t$, the closer to death or failure. 

For estimating the probability of conflicts, collisions are system failures, and our evaluation is based on the records of daily interactions. Contrary to lifetime survival analysis which measures survival duration and \textit{longer} duration is closer to deaths, here conflict analysis measures proximity and \textit{shorter} proximity is closer to collisions. The frequency of conflicts increases when proximity reduces. We thus define a ``conflict function'' $F(s)=\Pr(S<s)$, which represents the cumulative probability that the interaction proximity $S$ is less than a certain $s$. Recall that in Section \ref{sec: proximity-characterised conflict hierarchy} we use $f_S(s|\theta;\phi)$ to describe the proximity distribution of $S$ in an interaction context $\theta$. Rewriting the function as $f_S(s;\phi)$ for convenience, given that $\phi=\mathbb{I}(\theta)$ and $\theta=\mathbb{I}^{-1}(\phi)$, then Equation \eqref{eq: proximity cdf} shows the conflict function in an interaction context $\theta$ where the conflict hierarchy is characterised by $\phi$.
\begin{equation}\label{eq: proximity cdf}
\begin{aligned}
    F(s;\phi) &= \Pr(S<s;\phi) = \int_0^s f_S(x;\phi)\mathrm{d}x
\end{aligned}
\end{equation}

The shorter the $s$, the more likely a collision is to occur, but the likelihood varies with conflict intensity. Aligned with the conflict hierarchy in Figure \ref{fig: safety_pyramid}, we define conflict intensity as the inverse of conflict frequency, i.e., a conflict at intensity $n$ occurs once per $n$ times in the same interaction context. Specifically, assuming an interaction where the proximity between road users is $s$; if $s$ is always smaller than the observed proximities for $n$ times interactions in the same situation, this interaction is a conflict with a frequency of $1/n$, and we consider it an extreme event of intensity $n$. According to extreme value theory, we can use $\left(\Pr(S\geq s;\phi)\right)^n$ to calculate the cumulative probability that a proximity $s$ is the minima in $n$ times of observations in the same interaction context. Therefore, for an interaction at proximity $s$ and in the context $\theta$ that is characterised by $\phi$, its probability of conflict is not a value, but a function regarding conflict intensity $n$. We denote this function by $C$ as shown in Equation \eqref{eq: conflict probability}.
\begin{equation}\label{eq: conflict probability}
\begin{aligned}
    p(c|s,\phi):&=C(n; s,\phi) = \left(1-F(s;\phi)\right)^n = \left(\int_s^\infty f_S(x;\phi)\mathrm{d}x\right)^n 
\end{aligned}
\end{equation}

\begin{figure}[htb]
    \centering
    \includegraphics[width=0.91\textwidth]{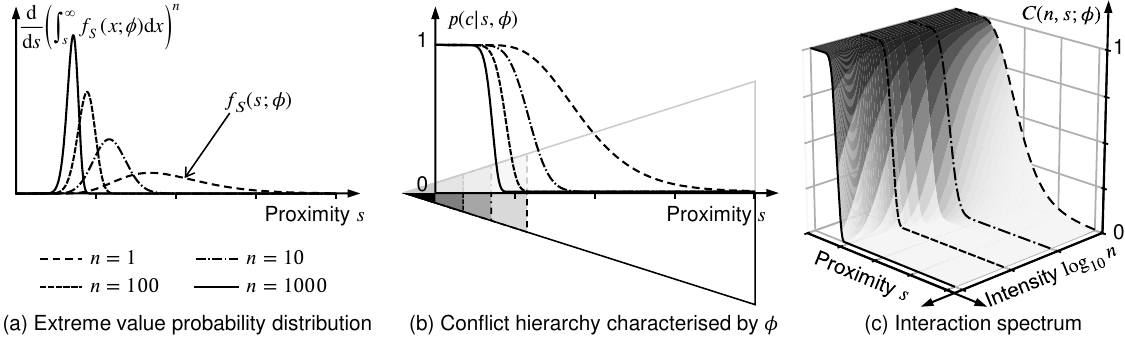}
    \caption{Interaction spectrum described with context-dependent proximity distribution and extreme value theory.}
    \label{fig: interaction_spectrum}
\end{figure}

Figure \ref{fig: interaction_spectrum}(a) shows the probability distributions of Equation \eqref{eq: conflict probability} under different intensity $n$. As displayed in Figure \ref{fig: interaction_spectrum}(b), this probability fulfils the requirements in Equations \eqref{eq: probability limit}, monotonically increasing while $s$ decreases. In Figure \ref{fig: interaction_spectrum}(c), we present an extreme value theory-based interaction spectrum described by Equation \eqref{eq: conflict probability}. It corresponds to the practical understanding of a conflict: the smaller the proximity and the lower the conflict intensity, the greater the conflict probability. Note that conflict probability does not equate to collision probability in our derivation. Collision probability specifically refers to the case when $n$ is very large and approaches infinity, representing the most intense and least frequent interaction. In that case, Equation \eqref{eq: conflict probability} converges to either Gumbel, Frechet, or Weibull distribution. Here we do not particularly consider this convergence at an infinite $n$, as traffic and driving safety focuses not just on collisions, but on conflicts at varying levels of intensity.

%//////////////////////
\section{Statistical learning tasks for conflict detection}\label{sec: learning}
\begin{figure}[htb]
    \centering
    \includegraphics[width=\textwidth]{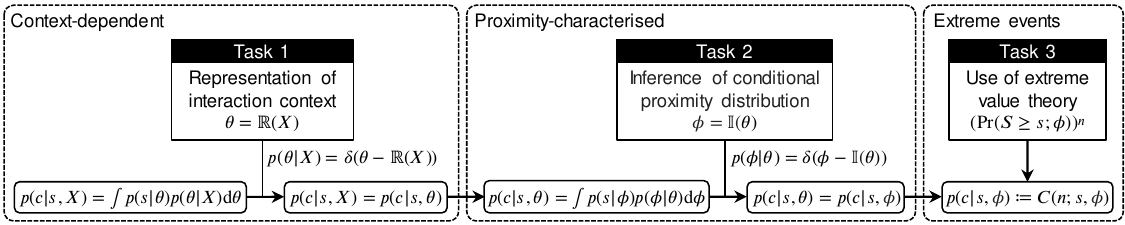}
    \caption{The unified framework and statistical learning tasks for conflict detection.}
    \label{fig: decomposition}
\end{figure}

%//////////////////////
The previous section explains the unified theoretical foundation of conflict detection. Now we continue by framing a series of statistical learning tasks for application in practice. Based on the derivation in Section \ref{sec: theory}, conflict detection estimates the probability of context-dependent and proximity-characterised extreme events, and we can decompose it into three tasks as shown in Figure \ref{fig: decomposition}. The first task $\theta=\mathbb{R}(X)$ maps situational observables $X$ to a compressed information space as $\theta$, which represents interaction context. In a certain interaction context $\theta$, the second task $\phi=\mathbb{I}(\theta)$ infers the conditional probability distribution of typical proximity behaviour of road users. Then the third task uses extreme value theory to determine the relations between proximity, conflict intensity, and conflict probability. When $\theta=\mathbb{R}(X)$ and $\phi=\mathbb{I}(\theta)$ are deterministic, we can rewrite Equation \eqref{eq: theory conflict detection} into Equation \eqref{eq: apply conflict detection}.
\begin{equation}\label{eq: apply conflict detection}
\begin{aligned}    
    p(c|s,X)=\iint p(c|s,\phi)p(\phi|\theta)p(\theta|X)\mathrm{d}\phi\mathrm{d}\theta=p(c|s,\phi):=C(n;s,\phi) \text{,}\\
    \text{if } p(\theta|X)=\delta(\theta-\mathbb{R}(X)) \text{ and } p(\phi|\theta)=\delta(\phi-\mathbb{I}(\theta))
\end{aligned}
\end{equation}

These three tasks can be integrated within a pipeline for end-to-end learning, and can also be performed separately. In the following subsections, we will explain the tasks further and present preliminary methods for application. We emphasise here that there are many useful learning methods, and the ones we use in this study are not necessarily the best. Future investigation on optimising suitable statistical learning techniques is required.

%---------------------
\subsection{Representation of interaction context}
%---------------------
The first task $\theta=\mathbb{R}(X)$ is to create an informative representation for interaction contexts by selecting and transforming relevant observables of interaction situations. Traditionally, key variables such as absolute speed, relative speed, and deceleration rate have been widely validated and used in existing surrogate measures of safety. However, there may be more factors at play than those hypothesised in the existing measures. As mentioned in Section \ref{sec: context-dependent collision risk}, $X$ may cover various aspects such as vehicle motion states, environment factors, and participant characteristics. Incorporating different factors can be particularly helpful for a comprehensive assessment of collision risk and for user-customised collision warnings.

If the observables of an interaction situation are numerous or use time-series and/or image formats, data-driven representation learning can be effectively employed. Such techniques can compress selected observables into a lower-dimensional space, where different interaction contexts are adequately encapsulated. One approach can be auto-encoders, of which the learning objective is to minimise the reconstruction error between compressed representation and original information. Another approach is contrastive learning, which aims to minimise the difference between the representations of similar samples while maximising the difference between dissimilar samples. Representation learning is an active and evolving field in machine learning, where many other methods are worth investigating. It is important to note that purely data-driven methods may not always capture the nuances of complex interaction situations. Therefore, integrating domain knowledge in traffic safety is necessary and can improve the robustness and reliability of context representation.

To efficiently present the theoretical contributions of this research, we do not perform data-driven representation learning in this study. However, we have some notes for readers on learning representations from variables relevant to motion states. Firstly, the output in the next task of inference is about the proximity between interacting road users. To avoid label leakage, we recommend not including a complete series of position or speed vectors for all involved road users. Secondly, we suggest transforming the variables into local coordinate systems. This transformation can be centred on every road user involved in an interaction situation, as the same situation may be perceived differently by different road users. Such view transformation can ensure the consistency and enhance the comparability of interaction context representations. Once transformed, an interaction context can be represented as instantaneous frames or over a continuous period.

%---------------------
\subsection{Inference of conflict hierarchy}
%---------------------
The second task $\phi=\mathbb{I}(\theta)$ needs to learn the conditional probability distribution of proximity in a certain interaction context, i.e., $p(s|\theta)$. Our previous study \cite{Jiao2023} used conditional sampling to infer this probability. Such sampling requires a large amount of data to ensure enough samples for each condition range, while the condition ranges are finite samples of an infinite space. In this study, we use Gaussian Process Regression (GPR) to avoid discretising the condition space, and the rest of this subsection will explain more details. Note that there are other useful methods to learn the conditional probability, but we choose GPR in order to obtain the equation of $p(s|\theta)$ and thus its analytical cumulative probability.

Many studies have found that lognormal distribution best fits the distribution of spatial and temporal gaps between road users \cite{Meng2012,Pawar2016,Anwari2023}. Therefore, we assume that the proximity $s$ in a certain interaction context $\theta$ follows a lognormal distribution. Recall that in Section \ref{sec: proximity-characterised conflict hierarchy} we use $f_S(s|\theta;\phi)$ to describe $p(s|\theta)$, and now we have $\phi=\{\mu,\sigma\}$ where $\mu$ and $\sigma$ parameterise a lognormal distribution as shown in Equation \eqref{eq: lognormal}. This implies that the logarithm of proximity, $\ln(s)$, follows a Gaussian distribution that is parameterised by the same $\mu$ and $\sigma$. 
\begin{equation}\label{eq: lognormal}
    f_S(s|\theta;\mu,\sigma)=\frac{1}{s\sigma\sqrt{2\pi}}\exp\left(-\frac{\left(\ln(s)-\mu\right)^2}{2\sigma^2}\right)
\end{equation}

The Gaussian distribution of $\ln(s)$ can be utilised to learn $p(s|\theta)$ with GPR. First, we consider a mapping $g:\theta\rightarrow \ln(s)$ between proximity $s$ and the interaction context $\theta$ that $s$ is in. Assuming a normally distributed noise in this mapping, we denote that $\ln(s)=g(\theta)+\epsilon$, where $\epsilon\sim\mathcal{N}(0,\sigma_\epsilon^2)$. Second, given that $\ln(s)$ is derived to be normally distributed, the distribution of function $g(\theta)$ is Gaussian. Then we can consider that $g(\theta)$ is drawn from a Gaussian Process, as shown in Equation \eqref{eq: q_GP}. This suggests that a $g(\theta)$ is one sample from the multivariate Gaussian distribution of all possible mappings. 
\begin{equation}\label{eq: q_GP}
   g(\theta)\sim \mathcal{GP}\left(m(\theta),K(\theta,\theta')\right),
\end{equation}
where $\mu=m(\theta)$ and $\sigma=K(\theta, \theta')$ are the mean function and covariance function that specify a Gaussian Process. For two similar interaction contexts $\theta$ and $\theta'$, $g(\theta)$ is expected to be close to $g(\theta')$. 

With GPR, the mean and covariance functions are learned from data. Considering that we may include many variables in the representation of interaction context and thus a fairly large amount of data is needed to serve training, we use the Scalable Variational Gaussian Process (SVGP) model with the python library GPyTorch \cite{gardner2018gpytorch}. As the cost to increase scalability, SVGP does not ensure an exact solution and may underestimate variance. To train this model effectively, we maximise a predictive log-likelihood in Equation \eqref{eq: training loss}, which is proposed in \cite{Jankowiak2020} and we adapt it here. In addition to the symbols that we have consistently used, $N$ is the number of samples; $u$ denotes inducing variables that are introduced to perform sparse training and reduce computational load; $q(u)$ is the variational distribution of $u$; $p(u)$ is the prior distribution of $u$; $D_{\text{KL}}\left[q(u)||p(u)\right]$ computes the Kullback–Leibler divergence of $q(u)$ and $p(u)$; and $\beta$ controls the regularisation effect of KL divergence.
\begin{equation}\label{eq: training loss}
\begin{aligned}
    L&=\mathbb{E}_{p_{\text{data}}(\ln(s),\theta)}\left[\ln\left(p(\ln(s)|\theta)\right)\right]-\beta D_{\text{KL}}\left[q(u)||p(u)\right]\\
    &\approx \sum_{i=1}^N \ln\left\{\mathbb{E}_{q(u)}\left[\int p(\ln(s_i)|g_i)p(g_i|u,\theta_i)\mathrm{d}g_i\right]\right\}-\beta D_{\text{KL}}\left[q(u)||p(u)\right]
\end{aligned}
\end{equation}

Theoretically, learning the mean and covariance functions can approximate any form of mapping $g$ between interaction context $\theta$ and proximity $s$. As explained in Sections \ref{sec: context-dependent collision risk} and \ref{sec: proximity-characterised conflict hierarchy}, a surrogate metric for traffic conflicts essentially assumes such a mapping, based on which a threshold is then determined to distinguish safe interactions and conflicts. Therefore, in theory, any metric based on spatial-temporal proximity is a particular case under our unified framework. Without assuming a specific format of $g$, the mapping embedded in data can be statistically derived. This suggests that, given the same interaction context, a data-driven metric using the unified probabilistic approach proposed in this study should be no less conflict-indicative than a pre-assumed metric. 

%---------------------
\subsection{Conflict probability estimation and intensity evaluation}
%---------------------
In the third task, we use extreme value theory to estimate conflict probability and evaluate conflict intensity. This task plays a similar role to the selection of thresholds when using traditional metrics of conflicts. After the previous two tasks, we can learn parameters $\phi=\{\mu,\sigma\}$ that characterise the conflict hierarchy in different interaction contexts represented with $\theta$. Based on the derivation in Section \ref{sec: EVT-based interaction spetrum}, then we can write conflict function $F(s;\phi)$ and obtain $C(n;s,\phi)$ that relates conflict intensity $n$ and conflict probability.

Given that we infer $f_S(s|\theta;\phi)$ as the probability density function of the lognormal distribution, the conflict function $F(s;\phi)$ in Equation \eqref{eq: proximity cdf} can be further derived as Equation \eqref{eq: lognormal conflict function}. Here $\erf(z)=2\int_{0}^z e^{-x^2}\mathrm{d}x/\sqrt{\pi}$ and is the Gaussian error function. Then conflict probability function is as shown in Equation \eqref{eq: lognormal conflict probability}. 
\begin{equation}\label{eq: lognormal conflict function}
\begin{aligned}
    F(s;\mu,\sigma) &= \int_0^s \frac{1}{x\sigma\sqrt{2\pi}}\exp\left(-\frac{\left(\ln(x)-\mu\right)^2}{2\sigma^2}\right)\mathrm{d}x = \frac{1}{2}+\frac{1}{2}\erf\left(\frac{\ln(s)-\mu}{\sigma\sqrt{2}}\right)
\end{aligned}
\end{equation}

\begin{equation}\label{eq: lognormal conflict probability}
\begin{aligned}
    C(n;s,\mu,\sigma) &= \left(1-F(s;\mu,\sigma)\right)^n = \left(\frac{1}{2}-\frac{1}{2}\erf\left(\frac{\ln(s)-\mu}{\sigma\sqrt{2}}\right)\right)^n
\end{aligned}
\end{equation}

There are two general purposes for conflict detection. The first is conflict probability estimation. For an interaction, given current proximity $s$ and parameters $\phi$ that characterise the conflict hierarchy in the interaction context, we can use Equation \eqref{eq: prob estimation} to estimate the probability of a potential collision at different levels of conflict intensity $n$. The second purpose is safety/conflict evaluation. Assuming that the probability of a potential collision is larger than the probability of no collision, i.e., conflict probability is larger than 0.5, we can use Equation \eqref{eq: safety evaluation} to evaluate the maximum possible conflict intensity. 
\begin{subequations}
\begin{alignat}{3}
    \label{eq: prob estimation}
    \hat{p} &= C(n;s,\mu,\sigma) & &= \left(\frac{1}{2}-\frac{1}{2}\erf\left(\frac{\ln(s)-\mu}{\sigma\sqrt{2}}\right)\right)^n, \quad & n &\geq 1 \\
    \label{eq: safety evaluation}
    \hat{n} &= C^{-1}(p;s,\mu,\sigma) & &= \frac{\ln p}{\ln\left(\frac{1}{2}-\frac{1}{2}\erf\left(\frac{\ln(s)-\mu}{\sigma\sqrt{2}}\right)\right)}, \quad & 0.5 &< p < 1
\end{alignat}
\end{subequations}

Equations \eqref{eq: prob estimation} and \eqref{eq: safety evaluation} are useful in different practices. Conflict probability estimation can be used to issue collision warnings, which alert human drivers or driving assistance systems to prevent potential collisions. Conflict intensity evaluation can be used to identify varying levels of conflict cases in daily traffic. This helps to assess the impact of an infrastructure or traffic policies on traffic safety, and then make according improvements. 

%//////////////////////
\section{Demonstration}\label{sec: demonstration}
%//////////////////////
This section applies the proposed unified framework and statistical learning tasks on real-world trajectory data, to demonstrate the characteristics of this new conflict detection approach.  Table \ref{tab: experiments} presents an overview of our experiment design. In Section \ref{sec: data and exp}, we introduce the datasets used and experiment details. Then respectively in Sections \ref{sec: probability estimation} and \ref{sec: intensity evaluation}, we show the experiment results in conflict probability estimation and conflict intensity evaluation. For convenience, we use \textit{approach} to collectively refer to the unified probabilistic approach proposed in this study; and use the term \textit{unified metric} and its abbreviation \textit{Unified} to refer to the surrogate metric trained with our approach. It is also important to emphasise that these experiments are designed primarily for demonstration. Further exploration is expected in future research. 
\begin{table}[htb]
\centering
\caption{An overview of experiment design for performance demonstration.}
\label{tab: experiments}
\begin{tabular}{cccc}
\toprule
\textbf{Purpose} & \textbf{Training data} & \textbf{Application data} & \textbf{Characteristics:} this approach \\ \midrule
\begin{tabular}[c]{@{}c@{}}Conflict probability\\ estimation \\ $\hat{p}=C(n;s,\mu,\sigma)$\end{tabular} & \begin{tabular}[c]{@{}c@{}}Trajectories involving\\lane-changes\\in highD\end{tabular} & \begin{tabular}[c]{@{}c@{}}Near-crashes\\in 100-Car NDS\end{tabular} & \begin{tabular}[c]{@{}m{5.5cm}@{}} 1) is no less conflict-indicative than any of PSD, DRAC, and TTC;\\2) is generalisable across datasets.\end{tabular} \\
\cmidrule{1-4}
\begin{tabular}[c]{@{}c@{}}Conflict intensity\\evaluation\\$\hat{n}=C^{-1}(p;s,\mu,\sigma)$\end{tabular} & \begin{tabular}[c]{@{}c@{}}Trajectories involving\\lane-changes\\in highD\end{tabular} & \begin{tabular}[c]{@{}c@{}}Lane-changes\\ in highD\end{tabular} & \begin{tabular}[c]{@{}m{5.5cm}@{}} 1) covers more diverse conflicts during lane-changes than TTC;\\2) detects conflicts in a long-tailed distribution of intensity.\end{tabular} \\
\bottomrule
\end{tabular}
\end{table}

%---------------------
\subsection{Data and experiment details}\label{sec: data and exp}
%---------------------
We use two naturalistic trajectory datasets in this study. First is the highD dataset that was collected at 6 different locations on German highways using drones \cite{highd}. It includes detailed information about vehicle types, sizes, and movements, with a positioning error typically less than 10 cm. Considering that a significant part of driving on highways is independent without interactions with other vehicles, we select trajectories that involve lane-changes in highD. In addition to the high-quality drone-collected data for model training, a dataset of real-world conflicts is necessary for demonstration. The second dataset we use is from the 100-Car Naturalistic Driving Study (NDS), which is an instrumented-vehicle study conducted in the U.S. over 2 years in the early 2000s \cite{100CarNDS}. An event database \cite{100CarData} resulted from the study compiles information on 68 crashes and 760 near-crashes. From the time-series sensor data of radars and accelerometers, we reconstruct bird's eye view trajectories for these events\footnote{We open-source our trajectory reconstruction code at \url{https://github.com/Yiru-Jiao/Reconstruct100CarNDSData}}. Due to missing values, inaccuracy of sensing, and the lack of ground truth, not all of the events can be reconstructed. We end up matching 180 events based on the constraint of insufficient space (distance less than 4.5 m) for undetected vehicles, including 11 crashes and 169 near-crashes as summarised in Table \ref{tab: events in 100Car data}.
\begin{table}[htb]
\centering
\caption{Summary of matched and selected events in 100-Car NDS data.}
\label{tab: events in 100Car data}
\begin{tabular}{lccc}
\toprule
\textbf{ Event happened with} & \begin{tabular}[c]{@{}c@{}} \textbf{Matched} \\ \textbf{crashes} \end{tabular} & \begin{tabular}[c]{@{}c@{}} \textbf{Matched} \\ \textbf{near-crashes} \end{tabular} & \begin{tabular}[c]{@{}c@{}} \textbf{Selected} \\ \textbf{near-crashes} \end{tabular} \\ \midrule
leading vehicle & 5 & 119 & 47 \\
following vehicle & 6 & 30 & 17 \\
vehicle in adjacent lane & 0 & 13 & 2 \\
vehicle turning across oncoming traffic & 0 & 4 & 0\\
vehicle crossing through an intersection & 0 & 2 & 0\\
vehicle in oncoming traffic & 0 & 1 & 0\\
\midrule
In total & 11 & 169 & 66 \\
\bottomrule
\end{tabular}
\end{table}

The first experiment estimates conflict probability at each time moment for the events in 100-Car NDS data. We utilise the estimation to issue collision warnings, and compare warning effectiveness with using 3 other commonly used surrogate metrics. Effective collision warning maximises true positives while minimising false positives. To reserve safe interactions within an event, we select events with a duration of at least 6 seconds, no hard braking (acceleration larger than -1.5 m/s$^2$) in the first 3 seconds, and speeds larger than 3 m/s at the first time moment for both vehicles involved. After the selection, 4 crashes and 66 near-crashes remained. Considering the relatively low reliability of data in crashes and for a consistent comparison, we use only the selected near-crashes, which are also shown in Table \ref{tab: events in 100Car data}. The 3 broadly used metrics for comparison are Proportion of Stopping Distance (PSD), Deceleration Rate to Avoid a Crash (DRAC), and Time-to-Collision (TTC). Table \ref{tab: indicators} provides an overview of them, where $\Delta s$ is the distance between vehicle bounding boxes and $\Delta v$ is relative velocity. For PSD we use a braking rate of 5.5 m/s$^2$, as emergency braking is typically between 1.6 and 5.5 m/s$^2$ \cite{Fambro2000,Deligianni2017}. 
\begin{table}[htb]
\centering
\caption{Existing surrogate metrics of conflicts that are used for warning comparison.}
\label{tab: indicators}
\begin{tabular}{ccc>{\centering}m{3cm}cc}
\toprule
\textbf{Metric} & \textbf{Calculation} & \textbf{Note} &\textbf{ Reference(s)} & \textbf{Proximity $s$} & \textbf{Context $\theta$}\\ 
\midrule
PSD & $\displaystyle\frac{\Delta s}{v_\text{follower}^2/2/\text{dec}}$ & $\text{dec}=$5.5 m/s$^2$ & \cite{allen1978} & $\Delta s$ & $v_\text{follower}^2$ \\
\rule{0pt}{20pt} % Adjust this value to change the space between rows
DRAC & $\displaystyle \frac{\|\Delta v\|^2}{2\Delta s}$ & approaching only & \cite{Cooper1976} & $\Delta s$ & $\|\Delta v\|^2$ \\
\rule{0pt}{20pt} % Adjust this value to change the space between rows
TTC & $\displaystyle\frac{\Delta s}{\|\Delta v\|}$ & approaching only & \cite{Hayward1972,Hyden1987} & $\Delta s$ & $\|\Delta v\|$ \\ 
\bottomrule
\end{tabular}
\end{table}

The second experiment evaluates conflict intensity at each time moment for lane-change interactions in the highD dataset. We use lane-changes to demonstrate the applicability of our unified metric to two-dimensional conflicts, which remain challenging to be integratively indicated by traditional metrics. We identify lane changes using the lane references provided in highD data. Then we determine the start and end moments of a lane-change based on the time when the vehicle deviates 1/3 of its vehicle width from the centerline of its current lane. In this way, overtaking is considered as two consecutive lane-changes. Seeing the vehicle making a lane-change as an ego vehicle, the lane-change is potentially interactive if there is a vehicle in front or at rear of the ego vehicle in either the original lane or the target lane. Then we use both the unified metric and two-dimensional TTC (2D-TTC) to evaluate conflict intensity (if any) at each time moment. Here 2D-TTC follows the typical definition of TTC \cite[Fig. 2,][]{tarko2018}, assuming constant velocities for two approaching vehicles at the moment of evaluation\footnote{We open-sourced an algorithm for fast computation of two-dimensional TTC at \url{https://github.com/Yiru-Jiao/Two-Dimensional-Time-To-Collision}. This algorithm is used for all experiments in this study.}. The smaller the TTC value, the higher the conflict intensity; when TTC is infinite, no potential collision is expected to occur.

In both experiments, we use a single unified metric trained with the trajectories involving lane-changes in highD data. In this way, the first experiment can additionally demonstrate the generalisability of our approach; and the metric thresholds calibrated with real-world near-crashes can be used in the second experiment to distinguish conflicts. For a fair comparison with existing metrics, we avoid complex representations for interaction context $\theta$. Firstly, we include the information used in PSD, DRAC, and TTC, i.e., $v_\text{follower}^2$, $\|\Delta v\|^2$, and $\|\Delta v\|$ as analysed in Table \ref{tab: indicators}. Since the lead-follow relationship evolves during lane-changing, we use the squared speeds of both ego vehicles (which change lanes) and target vehicles (in the surrounding). Further, we consider the accelerations of ego vehicles \footnote{We do not include target vehicle accelerations because this information is lacking in 100-Car NDS data and cannot be reliably derived due to speed fluctuations prior to potential collisions. This lack also prevents comparison with the metric Modified Time to Collision (MTTC).}, heading directions of target vehicles relative to their ego vehicles, and lengths of both ego and target vehicles. In order to indicate two-dimensional conflicts, we define proximity $s$ utilising the two-dimensional spacing proposed in \cite{Jiao2023}. This spacing is denoted by $(x,y)$ in the real-time transformed relative coordinate systems, and we convert $(x,y)$ into $(\rho, s)$ into polar coordinate systems, where $\rho$ is the angle between $(x,y)$ and $(1,0)$ within a range of $[-\pi,\pi]$. Then we add $\rho$ as one more variable in $\theta$, and use $s$ as the proximity measure to reflect conflict hierarchy. 

\begin{figure}[htb]
    \centering
    \includegraphics[width=\textwidth]{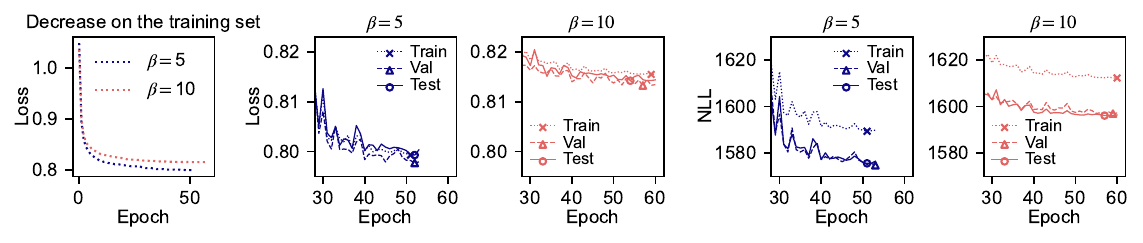}
    \caption{Evaluation of SVGP training and selection of the model to apply.}
    \label{fig: model_evaluation}
\end{figure}

We train two sets of SVGP models under different settings of the hyper-parameter $\beta$ in Equation \eqref{eq: training loss}, with $\beta=5$ and $\beta=10$. Our training is on 60\% of the selected interaction trajectories involving lane-changes; and the validation set and test set account for 20\% each. During the training, we reduce the learning rate dynamically to avoid overfitting of the models, and stop training early when validation loss converges. Figure \ref{fig: model_evaluation} shows the training progress, where there are 271 batches per epoch and 2,048 samples per batch. Training in both settings converges well, and stops earlier as well as reaches lower loss values when $\beta=5$. We then evaluate model performance by both loss and negative log-likelihood (NLL) on validation and test sets, as presented in Figure \ref{fig: model_evaluation}. As a result, we select the model achieving the minimum test loss and the second minimum test NLL, i.e., the model with $\beta=5$ after 52 epochs of training, to apply in the subsequent demonstration experiments.

Since we consider two-dimensional interactions, maintaining a consistent coordinate system is necessary. The highD dataset uses a coordinate system where the x-axis points from left to right and the y-axis points downwards. This is a mirrored system from the traditional engineering coordinate system where the y-axis points upwards. When applying a model that is trained on highD data to other datasets, such as the 100-Car dataset, it is essential to adjust for this inconsistency. We make adjustments by exchanging the x and y coordinates of positions, velocities, and heading directions in 100-Car NDS data before coordinate transformation. We hereby remind the readers to correct potential inconsistencies in the coordinate systems between training data and application data.

%---------------------
\subsection{Collision warning compared with existing metrics}\label{sec: probability estimation}
%---------------------
The first experiment based on conflict probability estimation is designed to demonstrate two characteristics of our approach. First, the unified metric performs at least as well as any among PSD, DRAC, and TTC in collision warning. This is determined in theory, but experimental results with real-world near-crashes can provide additional evidence. Second, we trained the unified metric with trajectories in the highD dataset while applying it to the events in 100-Car NDS data. This cross-validation involves not only the differences between German and American driving, but also the evolution of driving over more than 15 years. If the highD-trained metric performs well for 100-Car near-crashes, we can argue for the generalisability of this approach, which implies common principles in human road use interactions.

In order to systematically compare warning effectiveness, we define performance indicators as explained in Table \ref{tab: exp1_kpi}. Collision warning classifies safe and unsafe interactions based on the values and specific thresholds of metrics. Seeing the metrics as different classification models, we can compare their ROC curves, and a metric is better if the area under its curve is larger. For each metric, we then select an optimal threshold such that the corresponding point on their ROC curve is closest to the ideal point with zero false positive and all true positives (0\%, 100\%). Based on the optimal threshold, we further calculate warning periods and warning timeliness. The warning period of an event should ideally be close to 100\%, but good warning timeliness does not necessarily mean early warning. On average, people need 1 to 1.3 seconds in response to an obstacle by braking, and in emergencies this can be less than 1 second \cite{Summala2000,Markkula2016}. Therefore, good warning timeliness should not be too large, as it may distract people earlier than they need to be warned; neither should it be too small, as the best timing to prevent a potential collision may be missed.
\begin{table}[htb]
\centering
\caption{Definitions of the performance indicators for collision warning effectiveness comparison.}
\label{tab: exp1_kpi}
\begin{tabular}{lm{11.5cm}}
\toprule
\textbf{Performance indicator} & \textbf{Definition} \\
\midrule
True positive & Seeing the moment when conflicting vehicles reach the minimum distance as a critical moment, the 3 seconds prior to the moment are supposed to be dangerous; any warning in this period marks a true positive. \\ \cmidrule{1-2}
False positive & The first 3 seconds in each of the selected events are supposed to be safe, so any warning in this period marks a false positive. \\ \cmidrule{1-2}
\begin{tabular}[c]{@{}l@{}}True positive rate and\\ false positive rate (\%) \end{tabular} & The rate of true positives and false positives, respectively, among all events. True positive rate ideally approximates 100\% and false positive rate 0\%. \\
\cmidrule{1-2}
ROC curve & Receiver Operating Characteristic curve that plots true positive rate against false positive rate at various thresholds.\\
\cmidrule{1-2}
Warning period (\%) & For each event with warnings issued under the optimal threshold, the percentage of warned time moments within the annotated conflict period. \\ \cmidrule{1-2}
Warning timeliness (s) & For each event with warnings issued under the optimal threshold, the time interval from the last safe-unsafe shift of warning until the critical moment.\\
\bottomrule
\end{tabular}
\end{table}

\begin{figure}[htb]
    \centering
    \includegraphics[width=\textwidth]{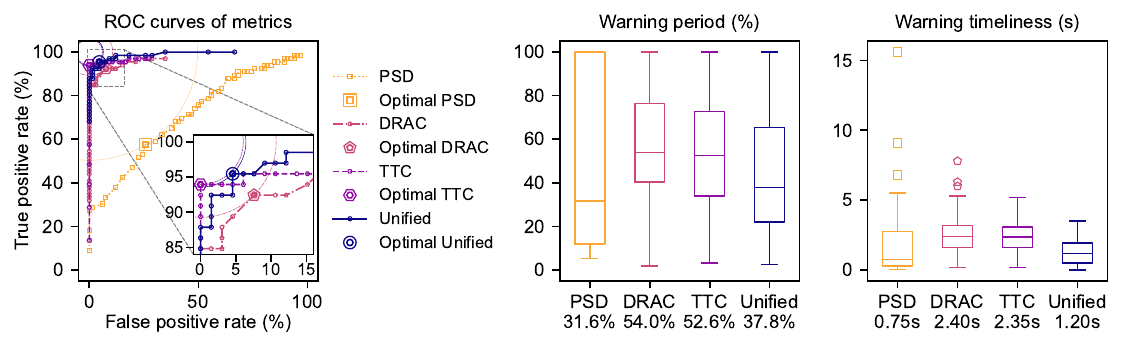}
    \caption{Collision warning effectiveness comparison between PSD, DRAC, TTC, and Unified. In the boxplots of warning period and timeliness, median values are marked below the labels of metrics.}
    \label{fig: warning_evaluation}
\end{figure}

Figure \ref{fig: warning_evaluation} presents the comparison of collision warning effectiveness. In the plot of ROC curves, we use circles centred at the ideal point (0\%, 100\%) and crossing the optimal points of different metrics to facilitate comparing their closeness to the ideal point. The metrics of DRAC, TTC, and Unified have comparable areas under their ROC curves, while PSD is less effective. Zooming in to see more details at the top left corner, TTC and Unified are consistently better than DRAC, and are very close to the ideal point. More specifically, the optimal threshold of Unified is $n^*=$17, reaching a true positive rate of 95.45\% and a false positive rate of 4.55\%; TTC$^*=$4.2 s and achieves 93.94\% true positive rate and 0.00\% false positive rate; DRAC$^*=$0.45, with a true positive rate of 92.42\% and a false positive rate of 7.58\%. Much less comparably, the optimal threshold of PSD is 0.52, with a true positive rate of 57.58\% only but a false positive rate of 25.76\%.

These metrics are further compared at their optimal thresholds by the box plots of warning period and warning timeliness in Figure \ref{fig: warning_evaluation}. Given the weak effectiveness of PSD, its warning period and timeliness are not comparable with other metrics. Both DRAC and TTC have higher median warning periods than Unified. This suggests that the unified metric may not issue a warning at the beginning of the annotated conflicts. This is also verified in the plot of warning timeliness, where DRAC and TTC have similar distributions and their median timeliness is both earlier than that of Unified. In addition, the warning timeliness of Unified is less varied. From a positive perspective, this means that using the unified metric gives consistent warnings and does not distract drivers too much; whereas from a negative perspective, this means that the unified metric may miss the chance to prevent a potential collision.

\begin{figure}[htbp]
    \centering
    \begin{subfigure}[b]{\textwidth}
        \centering
        \includegraphics[width=\textwidth]{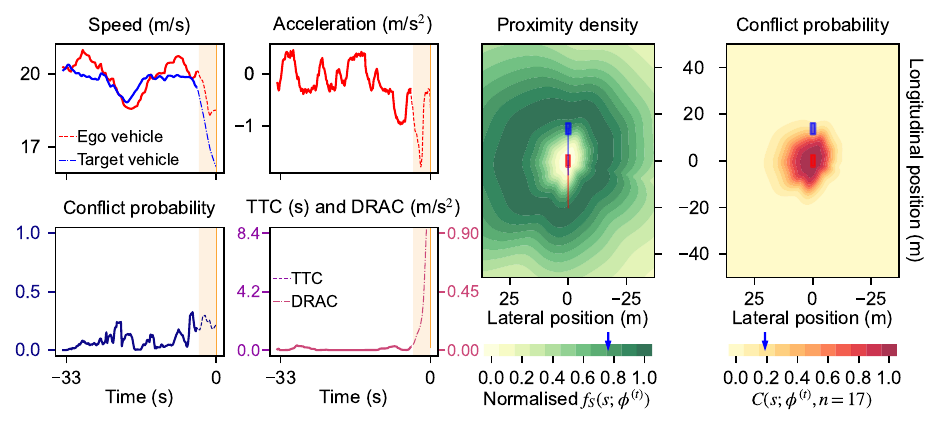}
        \caption{Frame 4.4 seconds prior to the critical moment of trip 8332.}
        \label{fig: ProbEst_8332_-440}
    \end{subfigure}
    \begin{subfigure}[b]{\textwidth}
        \centering
        \includegraphics[width=\textwidth]{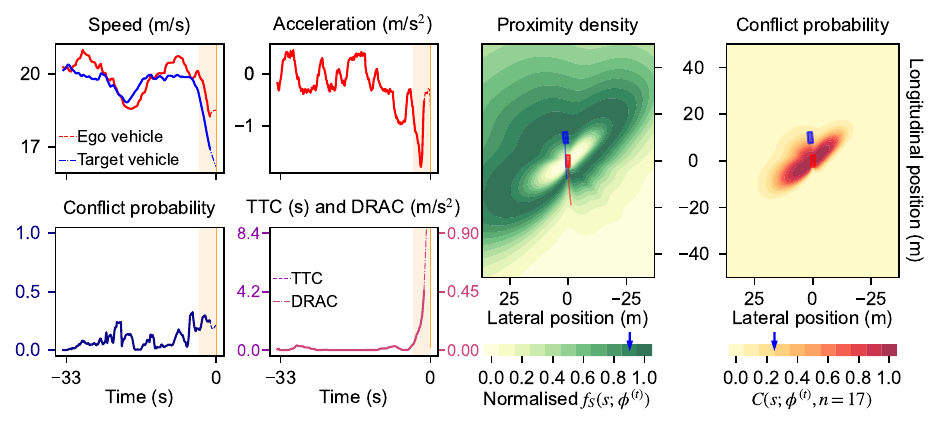}
        \caption{Frame 1.4 seconds prior to the critical moment of trip 8332.}
        \label{fig: ProbEst_8332_-140}
    \end{subfigure}
    \caption{Visualisation example for the collision warning of trip 8332.}
    \label{fig: prob_est_visualisation}
\end{figure}

For a more detailed analysis, Figure \ref{fig: prob_est_visualisation} shows two frames in the trip indexed by 8332. This event is warned by DRAC only, while TTC is larger than 10.59 s throughout and the conflict probability estimated by Unified remains lower than 0.5. In this figure, we present the profiles of speed, acceleration, and metric values, with the annotated conflict period shaded. We also plot heatmaps of proximity density distribution $f(s;\phi)$ and conflict probability function $C(s;\phi,n)$, where the y-axis points to the heading direction of ego vehicle and the x-axis points from right to left to align with highD's coordinate system. For each location around the ego vehicle, we assume the target vehicle is at that location and use actual interaction context $\theta$ for estimation. Figure \ref{fig: ProbEst_8332_-440} is the frame at 4.4 seconds before the critical moment; and Figure \ref{fig: ProbEst_8332_-140} is 1.4 seconds before. Notably, in Figure \ref{fig: ProbEst_8332_-140}, the proximity density distribution, i.e., probable positions of the target vehicle given the interaction context, does not point to the ego vehicle's heading direction. This is due to the lateral interaction between these two vehicles, which is in line with the narrative data of this event: ``the target vehicle was merging into the right lane ahead of the ego vehicle, causing the ego vehicle to brake to avoid a collision''. Comparing the two frames, the unified metric takes into account the ego vehicle's deceleration as successful prevention of an immediate conflict, and thus the estimated conflict probability stays low during the annotated conflict period.

We summarise the success and failure cases of conflict warnings in Table \ref{tab: warning results}. PSD is not included due to its lower effectiveness. For the events where at least one of Unified, TTC, and DRAC fails, as well as for the false warnings of Unified, we provide dynamic visualisations along with our open-sourced code. The readers are referred to the Acknowledgements to find links. Impressively, 58 out of the 61 near-crashes are correctly warned by all the metrics of Unified, TTC, and DRAC. This is because most events in the 100-Car dataset are rear-end conflicts during car following, for which TTC and DRAC have been proven to be useful. In the next experiment, TTC and our unified metric are challenged to handle two-dimensional traffic conflicts.
\begin{table}[htb]
\centering
\caption{Summary of success and failure cases of collision warnings by Unified, TTC, and DRAC.}
\label{tab: warning results}
\begin{tabular}{ccccl}
\toprule
\textbf{Unified} & \textbf{TTC} & \textbf{DRAC} & \textbf{Number of events} & \textbf{Trip ID} \\ \midrule
Succeed & Succeed & Succeed & 58 &  \\
Succeed & Succeed & Fail & 3 & 8622, 8854, 9101 \\
Succeed & Fail & Succeed & 0 & \\
Succeed & Fail & Fail & 2 & 8463, 8810 \\
Fail & Succeed & Succeed & 1 & 8761 \\
Fail & Succeed & Fail & 0 &  \\
Fail & Fail & Succeed & 2 & 8332, 8702 \\
Fail & Fail & Fail & 0 &  \\ 
\bottomrule
\end{tabular}
\end{table}

\newpage
%---------------------
\subsection{Conflict intensity evaluation for lane-changing interactions}\label{sec: intensity evaluation}
%---------------------
The second experiment evaluates the conflict intensity of lane-changing interactions, with which we want to demonstrate two more characteristics of our approach. First, this approach can cover a more diverse range of conflicts during lane-changes than TTC. This stands in theory because TTC assumes constant movements (velocities in this study) of interacting vehicles, which identifies potential lane-change conflicts only if two vehicles have crossing velocity directions and could collide without movement change. In contrast, a unified metric trained with our approach on two-dimensional daily driving data considers conflicts in all directions and varying interaction situations. Second, the intensity of conflicts evaluated by this approach is expected to have a long-tailed distribution. More specifically, the number of detected conflicts should decrease according to a power-law as conflict intensity increases. A long tail in such a distribution indicates that, although high-intensity conflicts are extremely rare, they occur with a non-negligible probability. Not all surrogate metrics of conflicts have this characteristic, but it is crucial to indicate very rare conflicts. 

\begin{figure}[htb]
    \centering
    \includegraphics[width=\textwidth]{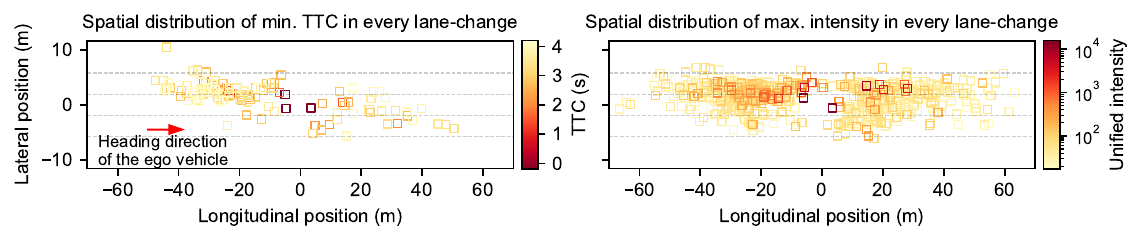}
    \caption{Spatial distributions of conflict moments with minimum TTC or maximum intensity in every lane-change. Target vehicle positions are transformed to a coordinate system centred at the ego vehicle position and with the longitudinal axis pointing to the ego vehicle heading direction. The dashed lines mark lanes at the average lane width in highD data.}
    \label{fig: highD_conflict_distribution}
\end{figure}

There are 13,364 lane-changes in the highD dataset, and in total 713 (5.34\%) of the lane-changes involve at least 1 second consecutively identified as a conflict by either TTC or the unified metric. This identification is under the optimal thresholds found in the first experiment, i.e., of 4.2 s for TTC and 17 for Unified. Among these lane-changes, in 679 the ego vehicle conflicts with one other vehicle, in 31 the ego vehicle conflicts with 2 other vehicles, and in 3 the ego vehicle conflicts with 3 other vehicles. These constitute 750 conflicts, among which 67 are indicated by both TTC and Unified; 18 are indicated by TTC only; and 665 are indicated by Unified only. In Figure \ref{fig: highD_conflict_distribution}, we plot the scattered relative locations of target vehicles to their ego vehicles aggregated at $(0,0)$. These locations are where a lane-changing interaction reaches the minimum TTC or the maximum intensity evaluated by Unified. 

Figure \ref{fig: highD_conflict_distribution} clearly shows that conflict detection by TTC strongly relies on the assumption of constant movements, while training a unified metric can cover all directions around the ego vehicle and accounts for the heterogeneity of proximity behaviour in every different direction. There are significantly more conflicts between the ego vehicles making lane-changes and the target vehicles in the left lane than those in the right lane. This may be due to more lane-changes from right to left or higher speeds of vehicles on the left lane, but future investigation is needed for more precise reasons. Notably, most of the lane-changing interactions have intensities lower than 100, which means one such conflict on average occurs in every 100 or fewer interactions. In comparison, conflicts with intensities higher than 100 are fewer, and those with intensities higher than 1000 are significantly fewer. However, despite their seeming rarity, they are far from unlikely to occur. In fact, foreseeing and preventing these safety-critical events remain key challenges for safe autonomous driving \cite{Liu2024}. 

\begin{figure}[htb]
    \centering
    \includegraphics[width=\textwidth]{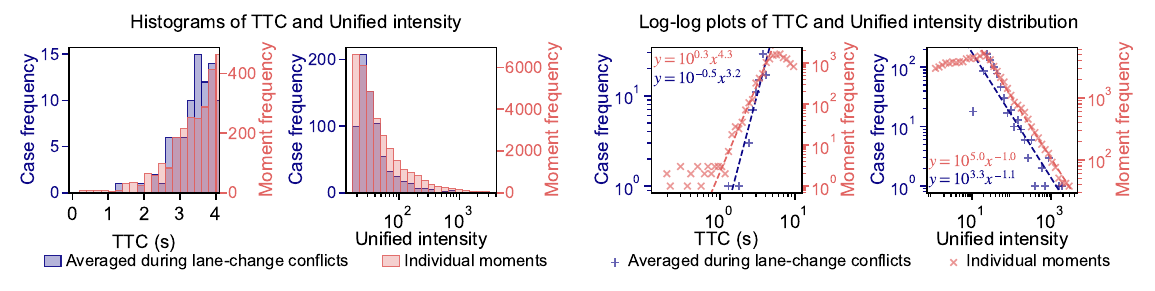}
    \caption{Intensity distributions during lane-changing interactions and at individual moments in the highD dataset.}
    \label{fig: conflict_intensity_hist}
\end{figure}

We present a closer look at the intensity distribution with Figure \ref{fig: conflict_intensity_hist}. In the left half of the figure, we plot the histograms of TTC values and Unified intensities respectively, for those at each individual moment and for averaged during each lane-change conflict case. The averaged values consider only lane-changing process, while the individual moments include car-following periods before and after lane-changes. The histograms show a decreasing frequency when intensity increases (TTC decreases), which is aligned with the assumption of conflict hierarchy. To further see if the distributions are long-tailed, in the right half of Figure \ref{fig: conflict_intensity_hist}, we make log-log scatter plots of the distributions, as well as their dashed trend lines and calibrated functions. All of the logarithm relationships are linear, therefore, all distributions to different extents have the characteristic of power-law. However, the trend line of TTC values averaged during lane-change conflicts has a higher slope than the line of TTC values at individual moments. This suggests that TTC detects fewer lane-changing conflicts than car-following conflicts. In contrast, the unified metric evaluates conflicts in the same theoretical framework, and thus has more consistent trend lines between averaging during lane-changes and at individual moments.

\begin{figure}[htp]
    \centering
    \begin{subfigure}[b]{\textwidth}
        \centering
        \includegraphics[width=\textwidth]{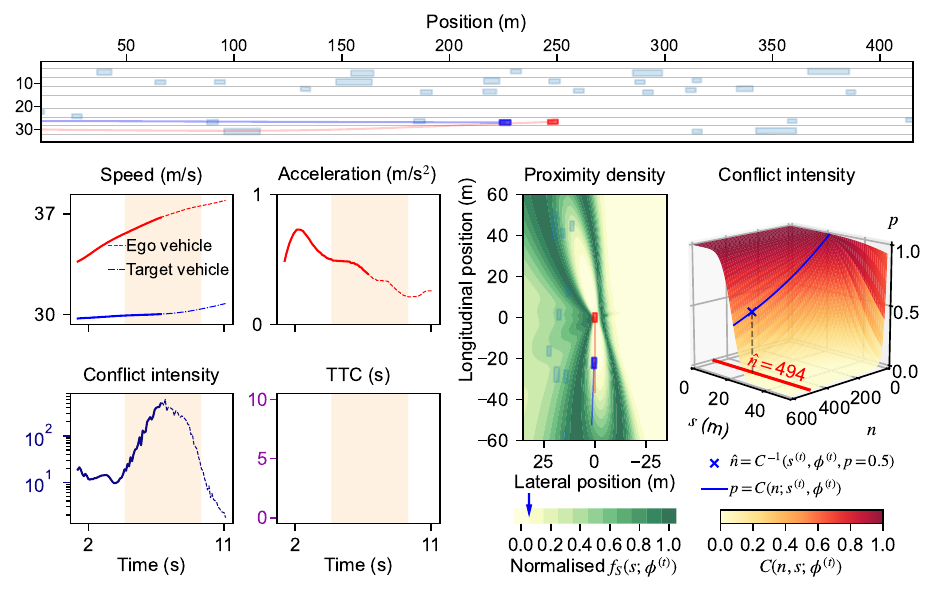}
        \caption{Frame 9719 at which the ego vehicle finishes first lane-change from the right lane to the middle lane.}
        \label{fig: IntEva_1_291860_291858_2909719}
    \end{subfigure}
    \begin{subfigure}[b]{\textwidth}
        \centering
        \includegraphics[width=\textwidth]{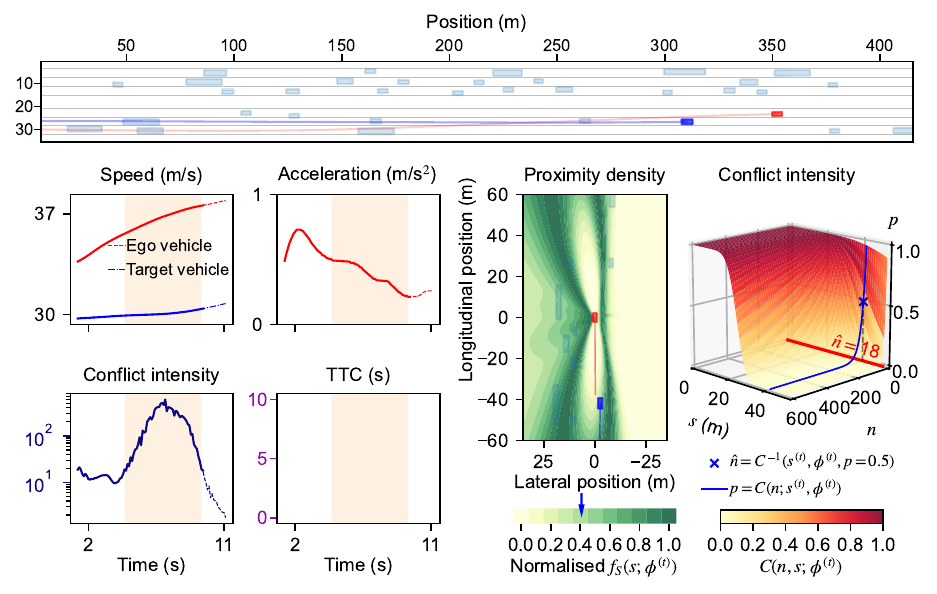}
        \caption{Frame 9747 at which the ego vehicle changes lane again to the left lane.}
        \label{fig: IntEva_1_291860_291858_2909747}
    \end{subfigure}
    \caption{Example dynamic visualisation of a lane-changing conflict detected by Unified only.}
    \label{fig: int_eva_visualisation}
\end{figure}

To provide more intuitive information, we make dynamic visualisations for the detected conflicts in different ranges of intensity indicated by TTC and Unified. A link can be found in the Acknowledgements. Here in Figure \ref{fig: int_eva_visualisation} we present an example at location 1 (indexed in the highD dataset) with vehicle track indices of 1860 and 1858. The ego vehicle in red makes two sequential lane-changes, and conflicts with the target vehicle in the intermediate lane. Similar to the visualisation for near-crashes in Section \ref{sec: probability estimation}, we plot the profiles of speed, acceleration, evaluated conflict intensity, and TTC values, as well as a real-time heatmap of proximity distribution. In the plots of profiles, lane-change periods are shaded. In addition, we visualise the interaction spectrum described by $C(n,s;\phi)$, which relates conflict intensity $n$, current proximity $s$, and conflict probability $p$.

The first frame in Figure \ref{fig: IntEva_1_291860_291858_2909719} is when the ego vehicle finishes its first lane-change from the right lane to the middle lane. In this frame, the conflict intensity is evaluated to reach the highest level in the whole process of sequential lane-change. On the surface of the interaction spectrum, we plot a line showing the relation between conflict intensity and probability at real-time proximity. This moment is at a very high probability to be considered as a minor conflict, and the maximum intensity to be considered as a conflict, i.e., the maximum intensity if conflict probability is larger than 0.5, reaches 494. This can be considered a serious conflict that occurs once per 494 times in the same interaction situations. Then the evaluated intensity decreases when the ego vehicle leaves the middle lane and continues moving to the left lane. In the frame shown in Figure \ref{fig: IntEva_1_291860_291858_2909747}, at the end of these two sequential lane changes, the evaluated intensity returns to safe levels.

\newpage
%//////////////////////
\section{Conclusion and discussion}\label{sec: conclusion}
%//////////////////////
Conflicts do not arise out of nowhere, and every conflict is an extreme continuation of preceding safe interactions. Based on this assumption, this study presents a unified probabilistic approach to detecting traffic conflicts. The unified framework of traffic conflict detection models conflicts as context-dependent extreme events in ordinary interactions. Under this framework, any existing surrogate safety measure that captures certain aspects of conflicts is a special case. Then the statistical learning tasks allow for data-driven hypotheses of traffic conflicts, making the unified detection at least as effective as using a pre-hypothesised surrogate measure. Preliminary experiments with real-world trajectory data demonstrate that the proposed approach provides effective collision warnings, generalises well across datasets, captures various conflicts consistently, and detects conflicts in a long-tailed distribution of intensity. 

These features enable consistent and comprehensive conflict detection, which can support scalable and reliable road safety research in the future. For example, more comprehensive surrogate measures of safety can be learned from naturalistic data. In the same way, an integrated estimation of collision risk becomes feasible for automated vehicles. Data-driven conflict detection also allows for the analysis of a broader range of factors underlying unsafe interactions in various situations. When the risk of conflicts is consistently evaluated across interaction contexts, it becomes possible to compare cross-modal road user interactions and provide consistent safety assessment of road infrastructures. Furthermore, safety-critical events that are rarely observed in data may be effectively identified, reinforced, and generated, contributing to training and testing automated vehicles. Pressing societal concerns, such as the impact of vehicle automation on road safety, could also be explored.

The current study has several limitations that should be investigated in future research. First, the demonstration in this study uses spatial proximity and trajectories in car-following and lane-changing scenarios; however, verification in more diverse traffic environments is needed. The framework could also be applied to temporal proximity, and involve other road users such as pedestrians and cyclists. Future research on cross-modal interaction safety is thus promising. Second, the proposed approach does not directly predict when a collision will occur. This could be addressed by incorporating trajectory prediction models and forming a more complete methodology for safe interaction planning. Third, the theoretical framework is based on proximity and does not indicate the severity of a potential collision. Collision severity depends on the energy released by a collision, which is more physics driven than extreme value theory-based. As severity is a key aspect of collision avoidance, further exploration is necessary. Lastly, the uncertainty in conflict probability estimation, which could inform the reliability of detection, is not quantified in this study. This quantification can be achieved by inferring the distributions of proximity characteristic parameters, given that the parameters are learned using Gaussian models. 

An important consideration regarding data should be noted. Although learning the representation of interaction context is identified as a key task, this study does not conduct representation learning. In theory, the use of learning methods can include almost all the information one can collect to describe the interactions between road users. This can go beyond movements and include road layouts, weather, individual characteristics of road users, etc. It is necessary to note, however, that the more information gets involved, the more diverse data is required to train an effective unified metric. Since the framework is well-suited for learning from accumulative evidence, future research could expand into continual learning as more data becomes available, allowing for increasingly robust and adaptive detection of traffic conflicts over time.

%////////////////////
\section*{Acknowledgments}
This work is supported by the TU Delft AI Labs programme. We thank Dr. Haneen Farah for her insightful discussion and internal review. We thank Dr. Guopeng Li for his discussion about technical details. We also would like to thank the editor and anonymous reviewers for their valuable comments and advice. 

We open-source our code for applying the approach and repeating our experiments at \url{https://github.com/Yiru-Jiao/UnifiedConflictDetection}. Dynamic visualisations of near-crashes in 100-Car NDS data are in the subfolder \href{https://github.com/Yiru-Jiao/UnifiedConflictDetection/tree/main/Data/DynamicFigures/ProbabilityEstimation/gifs}{./Data/DynamicFigures/ProbabilityEstimation/gifs/}; of lane-changing conflicts in highD data in the subfolder \href{https://github.com/Yiru-Jiao/UnifiedConflictDetection/tree/main/Data/DynamicFigures/IntensityEvaluation/gifs}{./Data/DynamicFigures/IntensityEvaluation/gifs/}.

%////////////////////
\section*{Declaration of generative AI and AI-assisted technologies in the writing process}
%////////////////////
During the preparation of this work the authors used ChatGPT in order to seek suggestions for readability improvement. No sentence was entirely generated by ChatGPT. After using this tool/service, the authors have reviewed and edited the content as needed and take full responsibility for the content of the publication.

%////////////////////
\newrefcontext[sorting=nyt]
\let\itshape\upshape
\printbibliography

% %////////////////////////////////////
% % \newpage
% \begin{appendices}
% \renewcommand\thefigure{\thesection.\arabic{figure}}
% \renewcommand{\theequation}{\thesection.\arabic{equation}}
% \section{Appendix}
% \setcounter{figure}{0}
% \renewcommand{\thefigure}{A\arabic{figure}}
% \setcounter{equation}{0}
% \renewcommand{\theequation}{A\arabic{equation}}
% \setcounter{table}{0}
% \renewcommand{\thetable}{A\arabic{table}}
% \setcounter{algocf}{0}
% \renewcommand{\thealgocf}{A\arabic{algocf}}
% %////////////////////////////////////
% %--------------------------------
% \subsection{Algorithm to identify lane-change}\label{sec:apdx_lc}
% %--------------------------------

% \end{appendices}

\end{document}